\documentclass[letterpaper]{article} 
\usepackage{aaai23}  
\usepackage{times}  
\usepackage{helvet}  
\usepackage{courier}  
\usepackage[hyphens]{url}  
\usepackage{graphicx} 
\def\UrlFont{\rm}  
\usepackage{natbib}  
\usepackage{caption} 
\usepackage{subcaption}
\usepackage{xcolor} 
\usepackage{algorithm}
\usepackage[noend]{algorithmic}

\usepackage{amsmath,amsfonts,bm}

\def\eqref#1{equation~\ref{#1}}

\def\1{\bm{1}}

\DeclareMathAlphabet{\mathsfit}{\encodingdefault}{\sfdefault}{m}{sl}
\SetMathAlphabet{\mathsfit}{bold}{\encodingdefault}{\sfdefault}{bx}{n}

\newcommand{\R}{\mathbb{R}}

\DeclareMathOperator*{\argmin}{arg\,min}

\usepackage{booktabs}       
\usepackage{multicol}
\usepackage{multirow}
\usepackage{makecell}
\usepackage{newfloat}
\usepackage{listings}
\DeclareCaptionStyle{ruled}{labelfont=normalfont,labelsep=colon,strut=off} 
\floatstyle{ruled}
\newfloat{listing}{tb}{lst}{}
\floatname{listing}{Listing}
\newcommand{\method}{XClusters}
\newboolean{techreport}
\setboolean{techreport}{true}

\title{\method{}: Explainability-first Clustering}
\author {
    Hyunseung Hwang,
    Steven Euijong Whang
}
\affiliations {
    KAIST\\
    \{aguno, swhang\}@kaist.ac.kr
}

\usepackage{bibentry}

\begin{document}

\maketitle

\begin{abstract}
We study the problem of {\em explainability-first clustering} where explainability becomes a first-class citizen for clustering.
Previous clustering approaches use decision trees for explanation, but only after the clustering is completed.
In contrast, our approach is to perform {\em clustering and decision tree training holistically} where the decision tree's performance and size also influence the clustering results.
We assume the attributes for clustering and explaining are distinct, although this is not necessary.
We observe that our problem is a monotonic optimization where the objective function is a difference of monotonic functions.
We then propose an efficient branch-and-bound algorithm for finding the best parameters that lead to a balance of cluster distortion and decision tree explainability.
Our experiments show that our method can improve the explainability of any clustering that fits in our framework.
\end{abstract}

\section{Introduction}

Explainable AI is becoming critical as AI is widely used in our everyday lives. A fundamental issue is that models are usually optimized for accuracy before being explained. As a result, significant effort is needed to make sense out of trained models, especially for complex ones. Even so, the explainability may not be sufficient to fully trust.

Instead, we contend that explainability must be a first-class citizen instead and focus on unsupervised learning. In particular, we propose the new problem {\em explainable-first clustering} where clustering is performed while balancing accuracy and explainability. As a motivating example, consider a temporal relational database that shows credit card spending trends for customers. In order to identify the key trends, a straightforward approach is to cluster the trends by Euclidean or Dynamic Time Warping (DTW)\,\cite{DBLP:conf/kdd/BerndtC94} distances. However, when explaining the clusters, each cluster may be a collection of a wide range of customer demographics, which makes it difficult to explain concisely. Instead, we would also like the clusters to be easy to describe as well.

More formally, we assume that the clustering is explained using decision trees, a problem that has attracted significant interest lately under the name of explainable clustering\,\cite{DBLP:conf/icml/MoshkovitzDRF20,DBLP:conf/icml/LaberM21,DBLP:conf/icml/MakarychevS21,DBLP:conf/nips/GamlathJPS21,DBLP:conf/aaai/BandyapadhyayFG22}. We consider smaller decision trees (i.e., they have fewer nodes) to be more explainable as they are easier to read\,\cite{lipton2018}. The features that are used for clustering are not necessarily the same as the ones for training the decision tree as in our motivating example. Our goal is to perform any clustering while ensuring that the decision tree trained on the clusters is as small as possible. There is a natural tradeoff between the cluster distortion and explainability, which we attempt to balance by solving an optimization problem. We call our method \method{} as it explicitly optimizes for explainable clusters.

We note that most explainable clustering approaches assume a fixed reference clustering where the clustering is performed {\em before} being explained\,\cite{DBLP:conf/icml/MoshkovitzDRF20,DBLP:conf/icml/LaberM21,DBLP:conf/icml/MakarychevS21,DBLP:conf/nips/GamlathJPS21}. Hence, the goal is to minimize the error of the decision tree that classifies examples to clusters. Even for a recent work that removes outliers from clusters for better explainability\,\cite{DBLP:conf/aaai/BandyapadhyayFG22}, the clustering is still done prior to outlier removal. 
In addition, if the data does not have many outliers, then removing data for the sake of explainability may result in incorrect clustering. 
While these approaches have the advantage of requiring little work on the clustering side, we contend that explainability must be incorporated in the clustering itself rather than using a single reference clustering for a truly explainable clustering. \method{} is thus {\em an orthogonal approach to existing explainable clustering techniques} where it can plug in any decision tree training within the novel holistic optimization of clustering and decision tree training.

\begin{figure}[t]
\centering
    \includegraphics[scale=0.5]{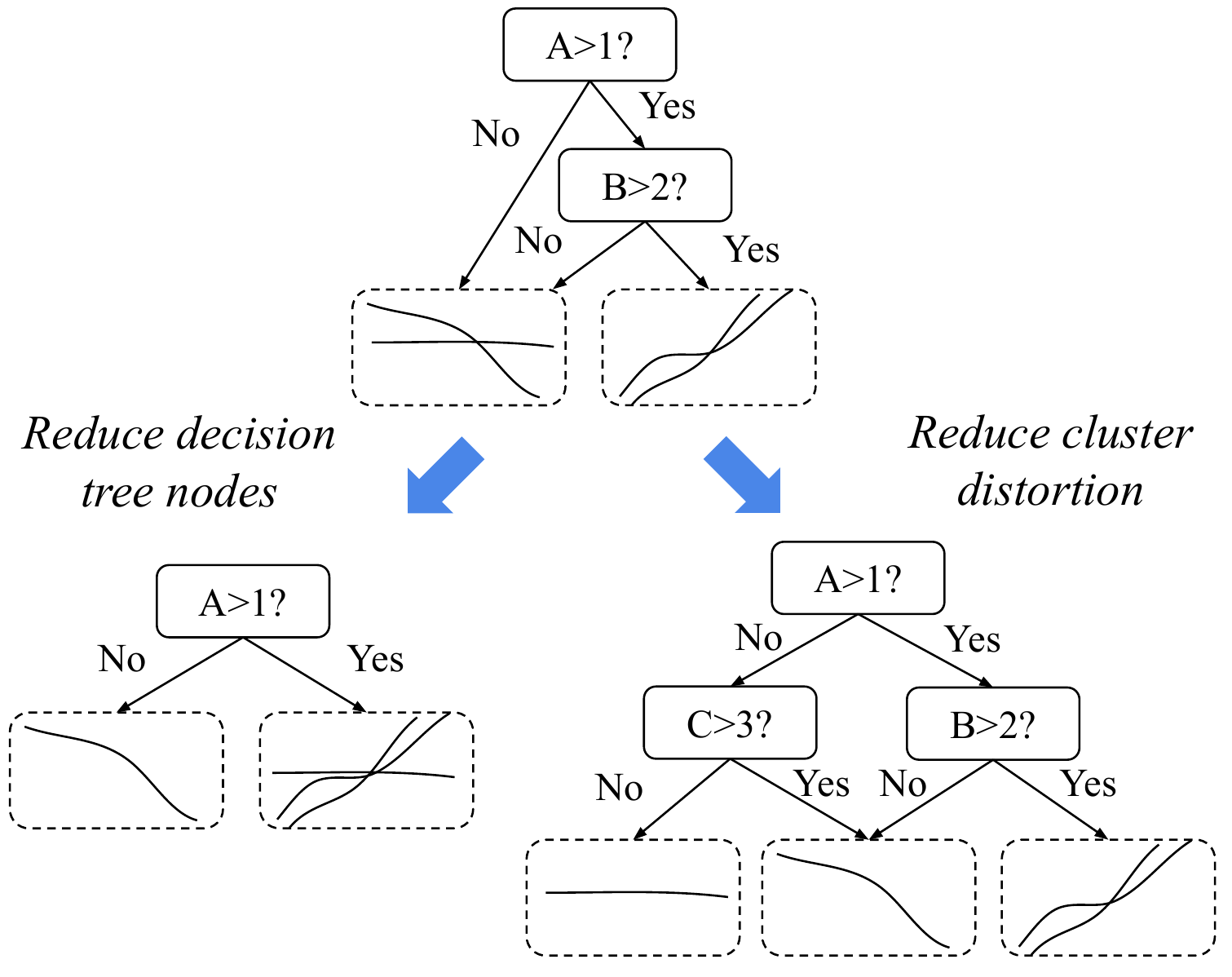}
    \caption{\method{}' strategy for training a decision tree during clustering. There are two objectives: reduce the cluster distortion for better clustering and reduce the decision tree nodes for better explainability. \method{} adjusts the number of clusters and ratio between accuracy and explainability features (explained later) to balance the objectives. } 
\label{fig:xclusters}
\end{figure}

Figure~\ref{fig:xclusters} illustrates how \method{} can trade off the cluster distortion for better explainability. Suppose that there are time-series trends that need to be clustered, but also explained using separate demographic features $A$, $B$, and $C$. Suppose that clustering by trend and then training a decision tree on top of the clusters results in the top figure. While most explainable clustering works take this approach, \method{} can further explore scenarios where the clustering also reflects demographic similarity and results in a smaller decision tree (bottom left) or the number of clusters is increased to reduce their distortion (bottom right). Notice that these two objectives affect each other and thus cannot be achieved separately. For example, reducing the cluster distortion results in a larger decision tree. The end goal is to minimize the decision tree's size and the distortion together. 

Our problem is a global optimization that is not convex. However, we make the interesting observation that the objective function is a difference-of-monotonic functions. We exploit this information and utilize existing monotonic optimization techniques\,\cite{DBLP:journals/tsp/MatthiesenHJU20,DBLP:journals/tsp/HellingsJRU12}, but tailor them to our problem setup. In particular, we propose an efficient branch-and-bound algorithm that tunes the parameters. 

In our experiments, we evaluate \method{} on various real time series relational datasets and show how it outperforms baselines that explain only after clustering or perform global optimization without exploiting monotonicity.

\section{Preliminaries}

Explainability has been studied extensively\,\cite{DBLP:conf/kdd/Ribeiro0G16}, but is fundamentally a subjective notion that depends on whether it is useful to users. Even for decision trees, the common view is that it is explainable, but others disagree as decision trees can be arbitrarily complex and large. In addition, even if there is some explainability, there are few proposals on how to quantify it. 

\paragraph{Explainability Measure}
We thus propose a concrete explainability measure for understanding clustering results. Suppose that there is a decision tree that is trained to classify examples to clusters. We measure the explainability of the decision tree using the classification accuracy and the size of the decision tree. A decision tree can only be explainable if it is accurate in the first place. In addition, the smaller the decision tree, the easier it is to read\,\cite{lipton2018}. If we assume an acceptable minimal accuracy of an explainable decision tree, then explainability can be measured as the size of the tree that has at least that accuracy. We measure the size of a decision tree by counting its nodes. Although other approaches count the number of levels, a tree that is shallow, but very bushy, may be difficult to understand.


\paragraph{Accuracy and Explainability Features}
We assume there are two groups of features: {\em accuracy features} and {\em explainability features}. The accuracy features are meant to maximize the clustering accuracy and thus lower cluster distortion. For example, if we are clustering time-series data, then the trends over time become the accuracy features. On the other hand, the explainability features are used to train a decision tree to classify examples to clusters for explanation. For example, if the time-series data is also relational with person attributes, then we can train a decision tree using the person attributes as the explainability features as illustrated in Table~\ref{tbl:aefeatures}. This type of data is common in finance (e.g., credit card companies or investment banks) or shopping (e.g., Amazon) industries where credit card transactions, stock values, and item purchases for certain demographics need to be analyzed over time. We assume that the two groups of features are not identical where clustering on the accuracy features leads to clusters with less distortion than when clustering on the explainability features for explaining purposes. It is still possible to use the same set of features for both accuracy and explainability, but then the distance functions applied on the features must be different in order to adjust the explainability during clustering. Our setup is not limited to time-series relational data and can be applied to various types of data including image data with metadata or any relational data.

\begin{table}[t]
  \caption{A sample credit card transaction table where each row contains a timestamp, transaction amount, and demographic features of the user. If we would like to analyze common spending trends and explain with demographics, then the time and amount columns become the accuracy features, and the others the explainability features.}
  \label{tbl:aefeatures}
  \centering
  \begin{tabular}{ccccc}
    \toprule
    \multicolumn{2}{c}{Accuracy Features} & \multicolumn{3}{c}{Explainability Features} \\
    \cmidrule(lr){1-2}\cmidrule(lr){3-5}
    Time & Amount & Age range & Zip code & Online \\
    \cmidrule(lr){1-2}\cmidrule(lr){3-5}
    0:50 & 50 & 10--20 & 456 & Yes \\
    2:00 & 300 & 30--40 & 123 & No \\
    2:30 & 200 & 50--60 & 999 & No \\
    \bottomrule
  \end{tabular}
\end{table}

\paragraph{Clustering}

We assume any clustering algorithm\,\cite{DBLP:journals/csur/JainMF99} that uses a distance function for clustering. We are thus not limited to a specific algorithm, although we do assume the clustering uses a distance function. As a default, we use DTW as the distance function as it is widely used and effective in identifying similar trends. We also use $k$-medoids clustering\,\cite{kaufman} as a default as it works naturally with DTW distances. When measuring distortion, we take the sum of squares of DTW distances from the examples to their closest center points. The center point of a cluster is its clustroid, which is the example with the lowest mean squared distances to the rest of the points in the cluster. Note that we cannot use centroids because we do not assume a Euclidean space.

For the distance function, there are two distances using the accuracy features ({\em a-distance}) and explainability features ({\em e-distance}). For example, one can use DTW for the trend distance and Jaccard distance for the explainability distance. 
A straightforward combination of the two measures is to normalize them and then take a weighted sum where $\alpha$ is a parameter used for the balancing:
\[\frac{(1-\alpha) \times \textrm{a-distance}}{\max\{\textrm{All\ a-distances\}}} + \frac{\alpha \times \textrm{e-distance}}{\max\{\textrm{All\ e-distances}\}} 
\]

\paragraph{Decision Tree Training}

We use decision trees to explain the clustering as they are widely considered more interpretable than other models. An alternative way to explain clusters is to use rule-based approaches. For example, a general boolean formula (GBF)\,\cite{DBLP:journals/pvldb/SinghMEMPQST17} is an if-then-else rule that can succinctly describe a cluster. \method{} can possibly be extended to GBFs because they can also be expressed as decision trees. 

Our framework is agnostic to the decision tree training algorithm. Although one can use recent techniques that attempt to optimize the decision tree to the clustering, it is also fine to use any other algorithm. What matters is the relative size of the decision trees, which we would like to minimize.

\section{Problem Definition}

We formulate our problem as minimizing two values: (1) $D$, which we define as the cluster distortion where a lower value is better as we would like the clusters to be coherent and (2) $N$, which we define as the number of decision tree nodes. Given that the decision tree's accuracy is sufficiently high, a smaller number of nodes is better as it means the decision tree is easy to read and thus explainable\,\cite{lipton2018}. 

\begin{equation}
\label{eq:problem}
\begin{split} 
& \min_{k, \alpha} \; D(k, \alpha) + \lambda N(k, \alpha) \\
\end{split}
\end{equation}
where we omit the normalization of $D$ and $N$ for brevity.

We use two parameters to adjust $D$ and $N$: $k$ is the number of clusters, and $\alpha$ is used to balance the accuracy and explainability distances as explained above. $k$ is suitable when using clustering algorithms like $k$-medoids, $k$-means, or hierarchical clustering, but one can use a different parameter depending on the clustering algorithm.

Without any assumptions, this global optimization problem is non-convex. In addition, the $D$ and $N$ objectives are not independent. For example, if we increase $k$ in order to decrease $D$, that also means $N$ is likely to increase due to more clusters. While one can use black-box optimization techniques like Bayesian Optimization\,\cite{DBLP:conf/ifip7/Mockus74}, we instead utilize monotonicity properties between the parameters and objectives for faster optimization as we explain from the next section.

\section{Monotonic Optimization}

We propose two practical monotonicity properties of the $D$ and $N$ objectives, which enable fast optimization. 

\begin{itemize}
    \item {\em As $k$ increases, $D$ is decreasing while $N$ is increasing.} As $k$ increases, there are more clusters, which means they tend to be smaller on average, causing the distortion to decrease. However, it becomes more difficult for a decision tree to precisely classify examples into the clusters.
    \item {\em As $\alpha$ increases, $D$ is increasing while $N$ is decreasing.} As $\alpha$ increases, the distance function is more about the explainability feature distance, so the clusters become less compact, causing the distortion to increase. On the other hand, the decision tree can be trained easier and thus requires fewer nodes to be sufficiently accurate.
\end{itemize}

In general, the monotonocity properties are not guaranteed to hold due to incomplete and imperfect data. For example, an increase in $k$ could suddenly cause one of the clusters to increase in size and thus increase the overall distortion. Or an increase in $\alpha$ may actually make the clusters more compact because the explainability features happen to be more effective than the accuracy features in reducing distortion.

However, our approach is inspired by learning lattice networks using partial monotonic functions\,\cite{DBLP:journals/jmlr/GuptaCPVCMME16,DBLP:conf/nips/YouDCPG17} where models are trained with monotonicity assumptions on model predictions against certain input features using domain knowledge. For example, if users are searching for restaurants on a website, it is reasonable to assume that they will more likely click on restaurants with higher ratings although there are exceptions where a user may want to explore new restaurants with lower ratings. Using this property is said to result in more accurate and flexible model training. Another analogy can be found in $k$-means clustering where an elbow method uses binary searching to find a $k$ value assuming that a larger $k$ results in a lower cluster distortion, even though there is no absolute guarantee of this trend. Likewise, our monotonicity properties can be thought as domain knowledge where given complete data, they should hold. In the worst case, \method{} will return a sub-optimal result while still being efficient. In our experiments on real datasets, we empirically show that the monotonicity properties mostly hold.

Using the monotonicity properties, we can re-formulate our problem into a monotonic optimization. Notice that increasing or decreasing $k$ and $\alpha$ have opposite affects on $D$ and $N$. We can thus rewrite the objective function of minimizing $D(k, \alpha) + \lambda N(k, \alpha)$ as minimizing $F(k, \alpha) - G(k, \alpha)$ where $F$ captures the decreasing $D$ and $\lambda N$ values when $\alpha$ and $k$ increase, respectively, and $G$ captures the decreasing $D$ and $\lambda N$ values when $\alpha$ and $k$ decrease, respectively. We cannot directly compute $F$ and $G$ as they are embedded in $D$ and $N$, but we only require their existence. The new formulation is then a difference of two monotonically 
increasing functions, which is a well-known optimization problem\,\cite{DBLP:reference/opt/Alizamir09}. Hence, our optimization problem can also be written as: 
\begin{equation}
\label{eq:problem}
\begin{split} 
& \min_{k, \alpha} \; F(k, \alpha) - G(k, \alpha) \\
\end{split}
\end{equation}
where $F$ and $G$ are decreasing functions of $k$ and $\alpha$.

\section{Methodology: \method{}}

We now solve the monotonic optimization problem efficiently.
There are largely two known methods for solving a difference-of-monotonic (DM) function problem: the Polyblock algorithm\,\cite{DBLP:reference/opt/Alizamir09,DBLP:journals/siamjo/Tuy00,tuy2001polyblock} and branch-and-bound algorithm\,\cite{DBLP:journals/siamjo/TuyMH06,tuy2005polynomial}.
The Polyblock algorithm is originally used to solve problems with monotonic objective functions and normal set constraints, which satisfies the condition that for any $x, y \in \R^n$, $x \in S$ and $y \leq x$ implies that $y \in S$ where $S$ is a normal set\,\cite{10.5555/1088007}. A DM problem can be converted to a monotonic optimization problem by introducing an auxiliary dimension. However, the Polyblock algorithm is known to be slow in practice\,\cite{DBLP:journals/tsp/HellingsJRU12}. Another popular method is the branch-reduce-and-bound (BRB) algorithm, which uses a branch-and-bound strategy to search parameters, but also has a reduce stage to improve convergence. The BRB algorithm is known to be faster than Polyblock and is thus more widely used.

We propose an efficient algorithm based on the branch-and-bound algorithm that is suitable for our problem (see Algorithm~\ref{alg:xclusters}). Our algorithm extends a branch-and-bound algorithm\,\cite{DBLP:journals/tsp/HellingsJRU12} that does not assume normal set constraints. The goal is to minimize the objective score $D + \lambda N$ by iteratively searching blocks of $k$ and $\alpha$ ranges. 

\begin{algorithm}[tb]
\caption{\method{} algorithm}
\label{alg:xclusters}
\textbf{Input}: training data $S$, maximum $k$ value $k_{\max}$\\
\textbf{Parameters}: $k$, $\alpha$\\
\textbf{Output}: clusters and decision tree
\begin{algorithmic}[1] 
\STATE $B \leftarrow [(1, 0), (k_{\max}, 1)]$
\STATE Compute upper and lower bounds of $B$
\STATE $B^* \leftarrow B$
\STATE $Q.push(B)$
\WHILE{$\neg Q.empty()$}
\STATE $B \leftarrow Q.pop()$ \text{// Block with lowest lower bound}
\IF {$B$'s normalized $k$ width is longer than the normalized $\alpha$ width}
\STATE $\{B_1, B_2\} \leftarrow$ Split $B$ by $k$ into two blocks
\ELSE
\STATE $\{B_1, B_2\} \leftarrow$ Split $B$ by $\alpha$ into two blocks
\ENDIF
\STATE Compute upper and lower bounds of $B_1$ and $B_2$
\STATE $Q.push(\{B_1, B_2\})$
\IF {$\min_{B \in Q} B.upper() < B^*.upper()$}
\STATE $B^* \leftarrow \argmin_{B \in Q} B.upper()$
\ENDIF
\STATE $Q \leftarrow Q \setminus \{B' \in Q | B'.lower() + \epsilon_b \geq B^*.upper()\}$
\ENDWHILE
\STATE \textbf{return} Clusters and decision tree of $B^*.upper()$
\end{algorithmic}
\end{algorithm}

For each block, we compute a lower and upper bound of the objective score. The conventional way to find a lower bound of a block is to subtract the $F$ value at the bottom-left point by the $G$ value at the top-right point exploiting the monotonicity properties. However, in our setting, we do not know the exact $F$ and $G$ functions and need to compute the lower bound using $D$ and $N$ only. Suppose that the bottom-left point is ($k_1$, $\alpha_1$), the top-right point ($k_2$, $\alpha_2$), and $k \in [k_1, k_2]$ and $\alpha \in [\alpha_1, \alpha_2]$. We know that
\begin{align*}
F(k, \alpha) - G(k, \alpha) &= D(k, \alpha) + \lambda N(k, \alpha) \\
& \geq D(k_2, \alpha) + \lambda N(k_1, \alpha) \\
& \geq D(k_2, \alpha_1) + \lambda N(k_1, \alpha_2).
\end{align*}
We thus use $D(k_2, \alpha_1) + \lambda N(k_1, \alpha_2)$ as the lower bound. Notice that we need to perform clustering and decision tree training to compute each of the $D$ and $N$ values, so minimizing the number of these trainings is important for efficiency. The upper bound can be computed by computing $D(k, \alpha) + \lambda N(k, \alpha)$ on an arbitrary point in the block. To save computation, we use the minimum of $D(k_2, \alpha_1) + \lambda N(k_2, \alpha_1)$ and $D(k_1, \alpha_2) + \lambda N(k_1, \alpha_2)$, whose components are already computed at this point to derive the lower bound. We avoid redundant clustering and decision tree training by keeping track of all ($k$, $\alpha$) results.

We iteratively split the block with the lowest lower bound score by its longer normalized width until there is no block to split. When splitting by $\alpha$, we simply divide the $\alpha$ range in half. When splitting by $k$, we split [$k_1,\ldots, k_n$] into [$k_1, \ldots, k_{\lfloor \frac{n}{2} \rfloor}$] and [$k_{\lfloor \frac{n}{2} \rfloor}, \ldots, k_n$]. Starting the second range with $k_{\lfloor \frac{n}{2} \rfloor}$ instead of $k_{\lfloor \frac{n}{2} \rfloor + 1}$ is intentional to save computation when computing lower and upper bounds of the split blocks. After generating two blocks $B_1$ and $B_2$, we check if they can be pruned. Notice that a block's upper bound is an actual objective score for some parameter values while the lower bound is potentially lower than the actual lowest objective score. We thus discard any block whose lower bound plus some tolerance $\epsilon_{b}$ is larger than or equal to any upper bound of another block. Using a higher $\epsilon_{b}$ results in more pruning at the cost of a sub-optimal result.

\begin{figure}[t]
\centering
    \includegraphics[scale=0.38]{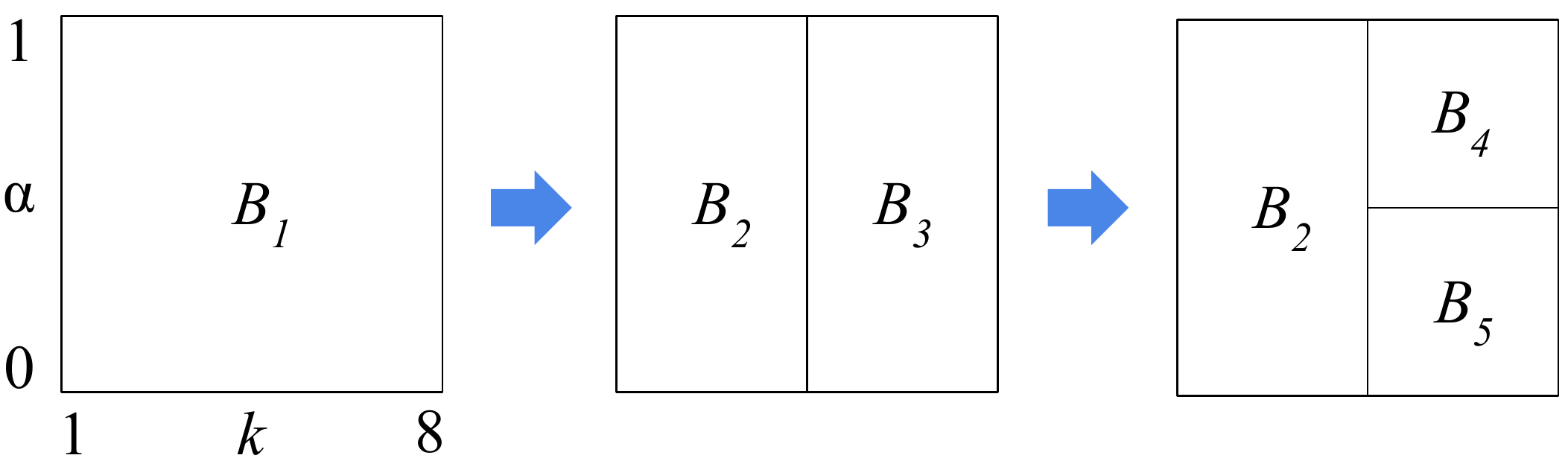}
    \caption{The \method{} algorithm iteratively splits blocks while pruning blocks that are not worth exploring based on their lower and upper bounds.}
\label{fig:algorithmexample}
\end{figure}

As a running example, suppose the algorithm starts with $k \in [1, 2, \ldots, 8]$ where $k_{max} = 8$, $\alpha \in [0, 0.1, 0.2, \ldots, 1]$, and $\epsilon_b = 0.1$ (see Figure~\ref{fig:algorithmexample}). The entire block is denoted as $B_1 = [(k = 1, \alpha = 0), (8, 1)]$. Suppose we split $B_1$ into $B_2 = [(1, 0), (4, 1)]$ and $B_3 = [(4, 0), (8, 1)]$, and they have lower and upper bound ranges of $[0.3, 0.4]$ and $[0.2, 0.5]$, respectively. Then, $B^* = B_2$ because $B_2$ has a smaller upper bound. We next split $B_3$ as it has a smaller lower bound. If the resulting blocks $B_4$ and $B_5$ have lower bounds at least $B^*.lower() + \epsilon_b = 0.4 + 0.1 = 0.5$, we can prune them as we know $B^*$ is a better block. We continue until $Q$ is empty and then return the clustering and decision tree of $B^*$'s upper bound result.

The complexity of Algorithm~\ref{alg:xclusters} to obtain an $\epsilon_b$-optimal solution can be derived using Theorem~4 in \,\cite{Vavasis1995} and is $O\left(\left(\frac{p}{\epsilon_b}\right)^{\frac{2}{q}}\right)$ assuming that $D + \lambda N$ is $q$-times differentiable, and the $q^{th}$ derivative is bounded by $p$. Here $q$ and $p$ can be viewed as constants that depend on the properties of $D + \lambda N$\,\cite{DBLP:journals/tsp/HellingsJRU12}. This result shows how fast \method{} optimizes when monotonicity holds perfectly. The less monotonicity holds, the worse \method{} performs.



\section{Experiments}

\paragraph{Datasets}

We use three real time series relational datasets. 
\begin{itemize}
    \item {\em Credit Card}: a proprietary dataset used in a major payment processing company. This dataset contains transactions of the credit card users before and after the COVID-19 outbreak (Dec. 2019 -- Jan. 2021) containing 5.7 billion transactions amounting to \$120B USD. The transactions include online or offline purchases made by each age group and gender for different business categories. We consider 2,551 demographics. We take a 90-day moving average of the time-series data.
    \item {\em DS4C}\,\cite{ds4c}: a public COVID-19 dataset containing patient data, policy data, and provincial data released by the Korea Centers for Disease Control \& Prevention (KCDC). We use floating population data of the city of Seoul for each age and gender group (Jan. 2020 -- May 2020). We take a 7-day moving average as the floating population data shows a cyclic pattern on a weekly basis. 
    \item {\em Contracts}\,\cite{contracts}: a public contract dataset maintained by the State of Washington. There are 170K contracts where each one contains customer, contract, vendor, and sales information. 
\end{itemize}

For the three datasets, the accuracy features are the time-series trends -- transaction amounts, population, and sales, respectively -- while the explainability features are the demographics information.

\paragraph{Measures}
\label{sssec:measures}

To evaluate a decision tree, we compute a weighted $F_1$ score on how it classifies examples to clusters. We weight each example in the Credit Card, DS4C, and Contracts dataset by its transaction amount, population, and sales amount, respectively. \ifthenelse{\boolean{techreport}}{In the appendix, we also use the Accuracy measure and obtain similar results.}{In our technical report\,\shortcite{techreport}, we also use the Accuracy measure and obtain similar results.}



\paragraph{Methods Compared}
\label{sssec:methodscompared}
We compare \method{} with baselines that represent the state-of-the-art approaches. 
\begin{itemize}
    \item {\em 2-Step}: Performs clustering and then trains a decision tree on the clusters. This approach represents previous works that assume fixed reference clusters where a decision tree can only be trained afterwards. We use the same clustering and decision tree training algorithms used for \method{} and set $\alpha = 0$. We also fix $k$ using an elbow method.
    \item {\em Grid Search (GS)}: Performs clustering and decision tree training using all the possible $k$ values considered by \method{} and a fixed set of $\alpha$ values in the range [$0, 0.05, 0.1, \ldots, 1$].
    \item {\em Bayesian Optimization (BO)}\,\cite{DBLP:conf/ifip7/Mockus74}: Performs clustering and decision tree training using Bayesian Optimization for tuning $k$ and $\alpha$. BO is widely used for hyperparameter tuning and has an O($n^3$) complexity where $n$ is the number of observations. We set the initial number of examples to explore and the number of iterations to be within the range [10, 30] and tune it to ensure a fair comparison with \method{}.
\end{itemize}

\paragraph{Other settings}
\label{othersettings}
We use Scikit-learn\,\cite{scikit-learn} for the decision tree training\ifthenelse{\boolean{techreport}}{ (we also evaluate with a recent explainable clustering technique\,\cite{DBLP:conf/icml/MoshkovitzDRF20} in the appendix).}{ (we also evaluate with \,\cite{DBLP:conf/icml/MoshkovitzDRF20} in our technical report\,\shortcite{techreport}).} For simplicity, we always make the decision tree overfit on the clusters and thus obtain perfect accuracy in classifying examples into clusters. We use the $k$-medoids algorithm\,\cite{kaufman} for clustering. Before performing any clustering, we compute the pairwise DTW distances between all example pairs. We search $k$ within the range [$3, 4, \ldots, 11$] for all datasets. For \method{}, we set $\lambda = 1$, and $\epsilon_b = 0.05$ as default values. We repeat each experiment 10 times. All experiments are performed on a server with Intel Xeon Gold 5115 CPUs.


\subsection{Monotonicity Properties}
\label{sssec:monotonicityproperties}
\begin{figure}[t]
\centering
    \subfloat[$k$ versus $D$]{
        {\includegraphics[scale=0.225]{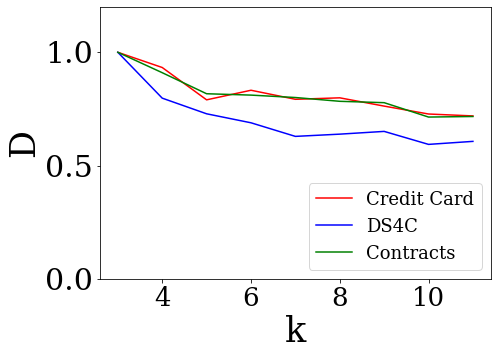}}
        \label{fig:kv}
        }
    \subfloat[$k$ versus $N$]{
        {\includegraphics[scale=0.225]{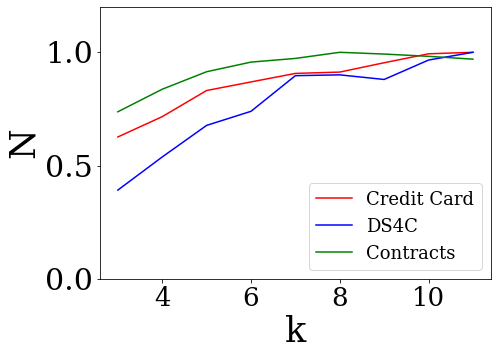}}
        \label{fig:kn}
        }\\
    \subfloat[$\alpha$ versus $D$]{
        {\includegraphics[scale=0.225]{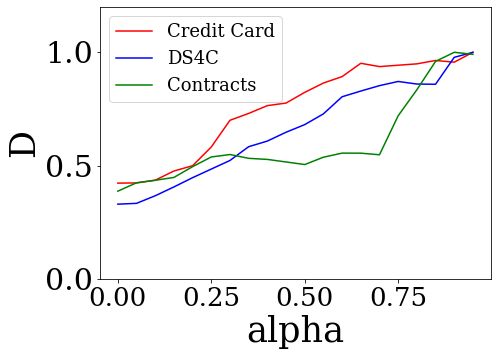}}
        \label{fig:kv}
        }
    \subfloat[$\alpha$ versus $N$]{
        {\includegraphics[scale=0.225]{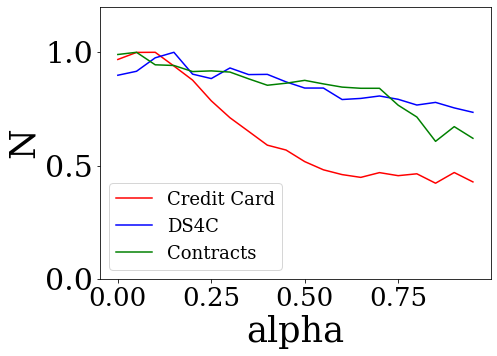}}
        \label{fig:kn}
        }
    \caption{Monotonic relationships between $k$ and $\alpha$, and the normalized objectives $D$ and $N$ for the three datasets.}
\label{fig:monotonic}
\end{figure}

We empirically verify the monotonocity assumptions, which form the basis of our algorithm. Figure~\ref{fig:monotonic} shows the trends between either $k$ or $\alpha$ and $D$ or $N$ for all the three datasets. For each $k$ ($\alpha$), we average the $D$ or $N$ values for all $\alpha$ ($k$) values considered by the GS baseline. As a result, all the trends are monotonic overall. However, some plots do have occasional violations due to incompleteness in the data. \ifthenelse{\boolean{techreport}}{In the appendix, we show similar monotonicity results using other distance measures.}{In our technical report\,\shortcite{techreport}, we show similar monotonicity results using other distance measures.}



\begin{table*}[t]
  \caption{Detailed comparison of \method{} and the three baselines -- 2-Step, GS, and BO -- using the default parameters. For each result, we show $D+ \lambda N$, $D$, $N$, and runtime in seconds. We repeat all experiments 10 times and take average values.}
  \label{tbl:accuracyexplainability}
  \centering
  \begin{tabular}{cccccc} 
    \toprule
     Dataset & Method & $D+ \lambda N$ & $D$ & $N$ & Runtime (sec) \\
    \midrule
    & 2-Step & $1.005_{\pm 0.000}$ & $0.166_{\pm 0.000}$ & $0.840_{\pm 0.000}$ & $0.426_{\pm 0.034}$\\
    Credit & GS & $0.625_{\pm 0.000}$ & $0.391_{\pm 0.000}$ & $0.234_{\pm 0.000}$ & $64.455_{\pm 0.506}$\\
    Card & BO & $0.657_{\pm 0.018}$ & $0.364_{\pm 0.039}$ & $0.293_{\pm 0.049}$ & $16.596_{\pm 0.377}$\\
    & \method{} & $0.656_{\pm 0.000}$ & $0.385_{\pm 0.000}$ & $0.271_{\pm 0.000}$ & $6.336_{\pm 0.104}$ \\
    \midrule
    \multirow{4}{*}{DS4C}  & 2-Step & $0.803_{\pm 0.000}$ & $0.101_{\pm 0.000}$ & $0.703_{\pm 0.000}$ & $0.009_{\pm 0.004}$\\
    & GS & $0.494_{\pm 0.000}$ & $0.193_{\pm 0.000}$ & $0.301_{\pm 0.000}$ & $1.359_{\pm 0.026}$\\
    & BO & $0.537_{\pm 0.078}$ & $0.126_{\pm 0.005}$ & $0.152_{\pm 0.034}$ & $1.554_{\pm 0.140}$\\
    & \method{} & $0.547_{\pm 0.000}$ & $0.216_{\pm 0.000}$ & $0.331_{\pm 0.000}$ & $0.122_{\pm 0.001}$\\
    \midrule
    \multirow{4}{*}{Contracts} & 2-Step & $0.916_{\pm 0.000}$ & $0.103_{\pm 0.000}$ & $0.814_{\pm 0.000}$ & $0.012_{\pm 0.001}$\\
    & GS & $0.644_{\pm 0.000}$ & $0.268_{\pm 0.000}$ & $0.375_{\pm 0.000}$ & $1.906_{\pm 0.044}$\\
    & BO & $0.650_{\pm 0.002}$ & $0.270_{\pm 0.004}$ & $0.381_{\pm 0.016}$ & $7.265_{\pm 0.562}$\\
    & \method{} & $0.644_{\pm 0.000}$ & $0.268_{\pm 0.000}$ & $0.375_{\pm 0.000}$ & $0.197_{\pm 0.015}$\\
    \bottomrule
  \end{tabular}
\end{table*}

\subsection{Explainability and Distortion Results}

Table~\ref{tbl:accuracyexplainability} compares \method{} with the other baselines in terms of $D + \lambda N$ and runtime. Recall that when running 2-Step, we need to fix $k$ using an elbow method by finding the smallest $k$ where the cluster distortion starts to converge. For the three datasets, the elbow point turns out to be $k = 5$ or $k = 6$. Since 2-Step cannot adjust the clustering for better explainability, it returns the worst results although it runs the fastest. GS usually finds the lowest $D + \lambda N$ with its brute-force searching, but does not outperform \method{} on the Contracts dataset because it only considers a fixed number of $\alpha$ values whereas \method{} explores an infinite space of values and can thus find better $\alpha$ values. When evaluating BO, we adjust its initial number of examples so that BO's $D + \lambda N$ becomes similar to that of \method{}. As a result, BO obtains a slightly lower $D + \lambda N$ than \method{} for the DS4C dataset, but is 12x slower. In comparison, \method{} exploits the monotonicity properties for fast searching. Moreover, BO is strictly worse than \method{} for the other datasets in terms of $D + \lambda N$ and runtime. While BO's runtime varies significantly depending on its number of initial examples, \method{} does not need such tuning to find near-optimal solutions. 



\begin{figure}[t]
\centering
 \subfloat[Varying $k$ and $\alpha$]{
    {\includegraphics[scale=0.225]{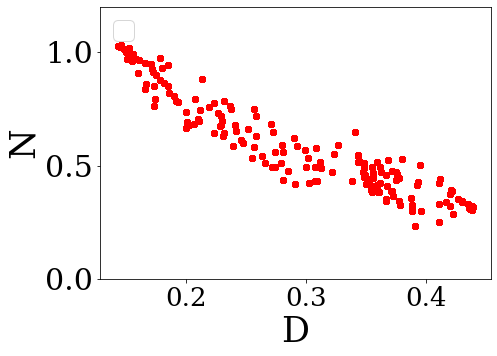}}
    \label{fig:ka}
 }
 \subfloat[Varying $\lambda$]{
    {\includegraphics[scale=0.225]{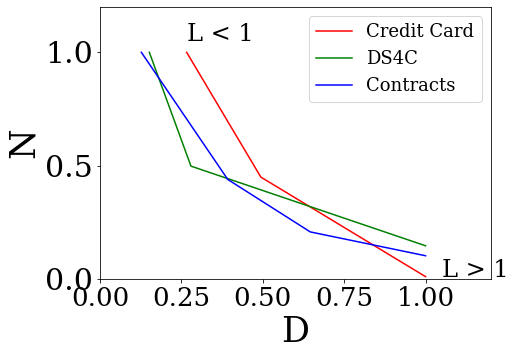}}
    \label{fig:lambda}
 }
 \caption{$D$ and $N$ tradeoffs on the datasets by varying the $k$, $\alpha$, and $\lambda$ parameters.}
\label{fig:tradeoff}
\end{figure}

\paragraph{Varying Parameters} We study the effect of varying $k$, $\alpha$, and $\lambda$ on the datasets. Figure~\ref{fig:ka} shows how $D$ and $N$ change for different $k$ and $\alpha$ values on the Credit Card dataset where the two objectives tradeoff as expected according to the monotonicity properties. Figure~\ref{fig:lambda} shows how $D$ and $N$ change against $\lambda$ when taking the results of GS with the lowest $D + \lambda N$ on the three datasets. A higher $\lambda$ means that there is more emphasis on explainability, and the decision trees tend to be smaller (lower $N$) with the risk of increasing the cluster distortions (higher $D$). 

\paragraph{Visualization} We visualize clusterings and their decision trees generated from the Credit Card dataset using Graphviz\,\cite{DBLP:conf/gd/EllsonGKNW00} in Figure~\ref{fig:treeandclusters}\ifthenelse{\boolean{techreport}}{ (see the appendix for larger images)}{ (see technical report~\shortcite{techreport} for larger images)}. We show two results with low $D + \lambda N$ scores: (a) $k = 5, \alpha = 0.45$ and (b) $k = 6, \alpha = 0.75$. The trees and clusters show how $D$ and $N$ can trade off. We note that some clusters of trends look quite noisy, but that is because the trends are clustered using DTW where they do not have to perfectly align visually to be clustered. Figure~\ref{fig:tc2}'s tree is smaller than Figure~\ref{fig:tc1}'s and is thus easier to understand. However, the clusters are relatively noisier where similar trends end up in different clusters, for the sake of a simpler decision tree. Which result is more desirable depends on the application and can be configured with $\lambda$.

\begin{figure}[t]
\centering
\subfloat[$k=5, \alpha=0.45$]{
    {\includegraphics[width=0.71\linewidth]{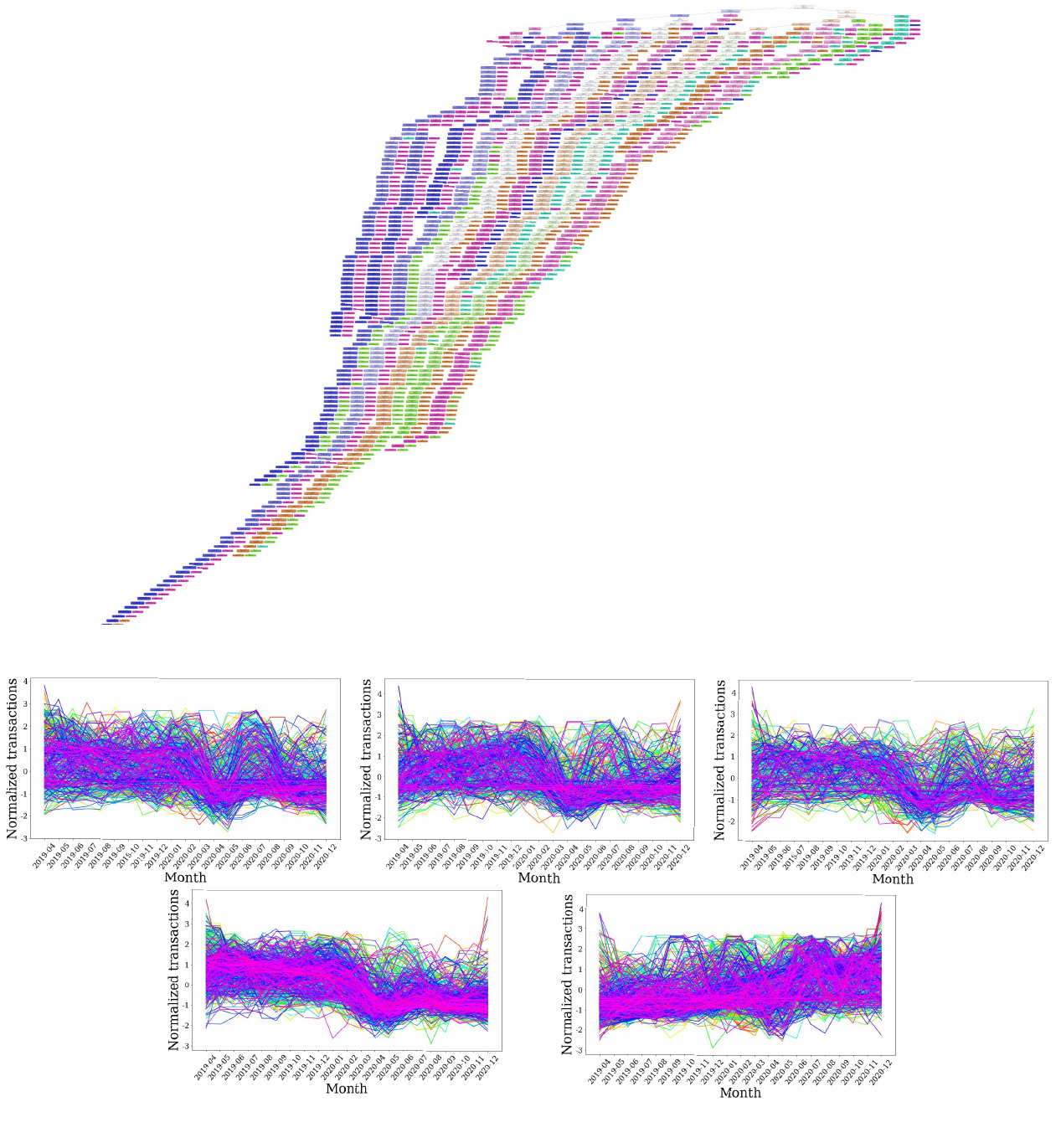}}
    \label{fig:tc1}
    }\\
\subfloat[$k=6, \alpha=0.75$]{
    {\includegraphics[width=0.71\linewidth]{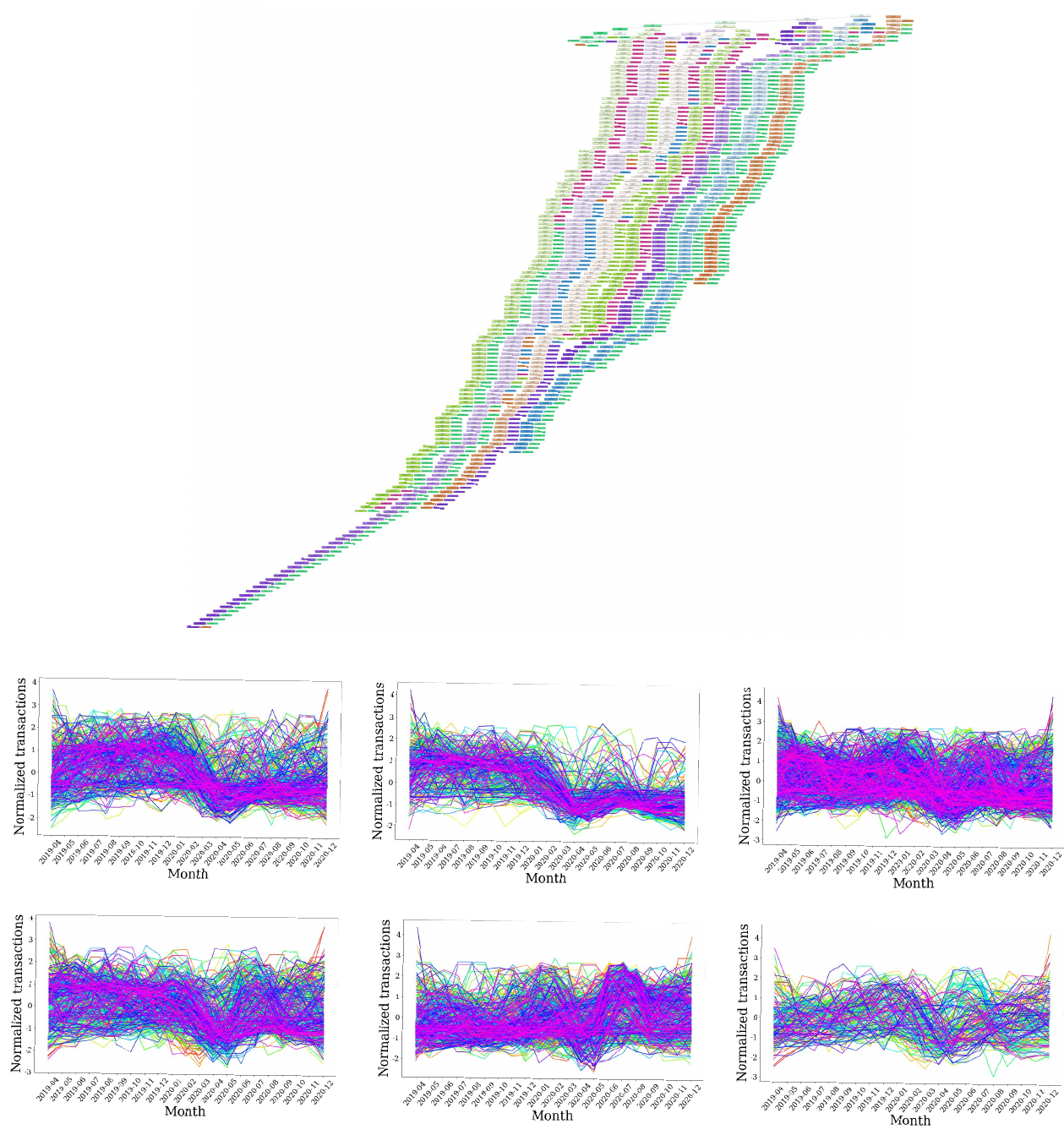}}
    \label{fig:tc2}
    }
\caption{\method{} results on the Credit Card dataset for two ($k$, $\alpha$) configurations that have similar low $D + \lambda N$ values. (a) and (b) have 1,703 and 1,477 nodes, respectively, and the visualizations demonstrate how $D$ and $N$ trade off.} 
\label{fig:treeandclusters}
\end{figure}

\subsection{Runtime Varying $\epsilon_b$}

We compare \method{}' runtime with the BO baseline while varying $\epsilon_b$, which determines how aggressively \method{} prunes blocks. Table~\ref{tbl:varyingepsilon} shows that \method{} can be significantly faster than BO on the three datasets while still having a lower or comparable $D + \lambda N$.

\begin{table}[t]
  \caption{Runtime comparison between \method{} and the BO baseline while varying $\epsilon_b$ on the three datasets.}
  \label{tbl:varyingepsilon}
  \centering
  \begin{tabular}{@{\hspace{4pt}}c@{\hspace{4pt}}c@{\hspace{8pt}}c@{\hspace{8pt}}c@{\hspace{8pt}}c@{\hspace{4pt}}}
    \toprule
    Dataset & Method & $\epsilon_b$ & $D+\lambda N$ & Runtime (sec)\\
    \midrule
    & \multirow{4}{*}{\method{}} & 0.01 & $0.625_{\pm 0.000}$ & $9.330_{\pm 0.019}$ \\     
    & & 0.05 & $0.656_{\pm 0.000}$ & $6.336_{\pm 0.104}$ \\
    Credit & & 0.10 & $0.656_{\pm 0.000}$ & $3.278_{\pm 0.017}$ \\
    Card & & 0.2 & $0.656_{\pm 0.000}$ & $2.119_{\pm 0.016}$ \\
    \cmidrule(l){2-5}
    & \multirow{1}{*}{BO} & n/a & $0.657_{\pm 0.018}$ & $16.596_{\pm 0.377}$ \\
    \midrule
    \multirow{5}{*}{DS4C} & \multirow{4}{*}{\method{}} & 0.01 & $0.547_{\pm 0.000}$ & $0.166_{\pm 0.023}$ \\     
    & & 0.05 & $0.547_{\pm 0.000}$ & $0.122_{\pm 0.001}$ \\
    & & 0.10 & $0.547_{\pm 0.000}$ & $0.081_{\pm 0.003}$ \\
    & & 0.2 & $0.547_{\pm 0.000}$ & $0.045_{\pm 0.002}$ \\
    \cmidrule(l){2-5}
    & \multirow{1}{*}{BO} & n/a & $0.537_{\pm 0.078}$ & $1.554_{\pm 0.140}$ \\ 
    \midrule
    \multirow{5}{*}{Contracts} & \multirow{4}{*}{\method{}} & 0.01 & $0.644_{\pm 0.000}$ & $0.207_{\pm 0.033}$ \\     
    & & 0.05 & $0.644_{\pm 0.000}$ & $0.197_{\pm 0.015}$ \\
    & & 0.10 & $0.644_{\pm 0.000}$ & $0.150_{\pm 0.013}$ \\
    & & 0.2 & $0.710_{\pm 0.000}$ & $0.065_{\pm 0.002}$ \\
    \cmidrule(l){2-5}
    & \multirow{1}{*}{BO} & n/a & $0.650_{\pm 0.002}$ & $7.265_{\pm 0.562}$ \\
    \bottomrule
  \end{tabular}
\end{table}

\section{Related Work}

Explainable AI is a broad field that attempts to explain trained models (see many surveys\,\cite{doshivelez2017rigorous,lipton2018,Tjoa2020,arrieta2020explainable,molnar2019}). There are many explaining techniques for classifiers in general\,\cite{DBLP:conf/kdd/Ribeiro0G16,DBLP:conf/aaai/Ribeiro0G18,DBLP:journals/jmlr/WangRDLKM17,DBLP:conf/kdd/LouCG12,DBLP:conf/nips/AdebayoGMGHK18}. Most of these techniques attempt to explain a model {\em after} the training is finished. In comparison, our focus is to make a model explainable during its training.

A recent line of work focuses on training decision trees on top of clustering for explanation. Explainable clustering\,\cite{DBLP:conf/icml/MoshkovitzDRF20,DBLP:journals/corr/abs-2006-02399,DBLP:conf/icml/LaberM21,DBLP:conf/icml/MakarychevS21,DBLP:conf/nips/GamlathJPS21} makes the $k$-means and $k$-medians algorithm results interpretable by showing a small decision tree that partitions the input data into clusters. ExCut\,\cite{DBLP:conf/semweb/Gad-ElrabSTAW20} explains embedding-based clustering results over knowledge graphs by also performing rule mining. More recently, there is an emphasis in making the explaining decision tree shallow\,\cite{DBLP:journals/corr/abs-2112-14718}. However, these methods assume a fixed reference clustering (e.g., a $k$-means clustering result) and fit a decision tree to that clustering. In comparison, \method{} varies the reference clustering to possibly find more explainable decision trees. Another key difference is that \method{} uses accuracy features for clustering and explainability features for decision tree training although the two types of features may overlap.



The most relevant work to \method{} is an outlier removal method that removes outliers from clusters to make a decision tree train accurately on them\,\cite{DBLP:conf/aaai/BandyapadhyayFG22}. Given a set of clusters, the objective is to find outlier examples to remove and a decision tree that exactly explains the clusters without the outliers. Here the explainability is measured as how many examples need to be removed for a perfectly-accurate decision tree. In contrast, we do not assume that some data is incorrect and can be removed. In addition, we do not assume the clusters are given as an input, but optimize the clustering itself for better explainability. An interesting future work is to remove outliers and optimize clustering for explainability together.

Monotonic optimization has been studied extensively\,\cite{DBLP:journals/siamjo/Tuy00,DBLP:reference/opt/Alizamir09,10.5555/1088007,tuy2001polyblock,DBLP:journals/siamjo/TuyMH06,tuy2005polynomial} where its techniques have impacted signal processing applications\,\cite{DBLP:journals/tsp/HellingsJRU12,DBLP:journals/tsp/MatthiesenHJU20}. The popular algorithm are the Polyblock and branch-reduce-and-bound (BRB) algorithm where the latter is known to be faster in practice. Our problem also falls into monotonic optimization, and we specialize the BRB algorithm to our setting. We note that our key contribution is identifying the monotonicity properties in our problem setup, and utilizing them for faster tuning.

\section{Conclusion}

To our knowledge, we are the first to propose an {\em explainable-first clustering} technique where explainability is also a primary objective of clustering. Given features for clustering and explaining, our \method{} framework minimizes the cluster distortion and the size of the decision tree trained to explain the clusters. We observed that our optimization is a difference of monotonic functions and proposed a branch-and-bound algorithm that efficiently finds the optimal number of clusters and distance function balancing ratio. We empirically showed on real datasets how \method{} outperforms baselines that explain after clustering or use black-box optimization without exploiting the monotonicity properties. A future work is to apply our techniques on various other types of data including images.

\section*{Acknowledgments}

This work was supported by the National Research Foundation of Korea(NRF) grant funded by the Korea government(MSIT) (No. NRF-2022R1A2C2004382). We appreciate BC Card for providing the Credit Card dataset. Steven E. Whang is the corresponding author.

\bibliography{main}

\ifthenelse{\boolean{techreport}}{
\clearpage
\onecolumn
\appendix
\section{Accuracy for Evaluating Clusters}
Continuing from the Experiments section, we repeat the experiments for Table~\ref{tbl:accuracyexplainability} when using the Accuracy measure (i.e., the portion of correct predictions among all predictions) instead of $F_1$ and compare with the same baselines. For XClusters, we set $\lambda=1$ and $\epsilon_b = 0.05$ as default values. We repeat each experiment 10 times. In Table~\ref{tbl:accuracyexplainability2}, we observe that the results are similar to Table~\ref{tbl:accuracyexplainability} except for the BO results due to the randomness of selecting initial examples. Hence, using the Accuracy measure produces similar results.

\begin{table*}[h]
  \caption{Comparison of \method{} and the three baselines using the Accuracy measure for evaluating decision trees.}
  \label{tbl:accuracyexplainability2}
  \centering
  \begin{tabular}{cccccc} 
    \toprule
     Dataset & Method & $D+ \lambda N$ & $D$ & $N$ & Runtime (sec) \\
    \midrule
    & 2-Step & $1.005_{\pm 0.000}$ & $0.166_{\pm 0.000}$ & $0.840_{\pm 0.000}$ & $0.399_{\pm 0.023}$\\
    Credit & GS & $0.625_{\pm 0.000}$ & $0.391_{\pm 0.000}$ & $0.234_{\pm 0.000}$ & $62.306_{\pm 0.271}$\\
    Card & BO & $0.650_{\pm 0.022}$ & $0.378_{\pm 0.016}$ & $0.271_{\pm 0.030}$ & $16.653_{\pm 0.638}$\\
    & \method{} & $0.656_{\pm 0.000}$ & $0.385_{\pm 0.000}$ & $0.271_{\pm 0.000}$ & $6.274_{\pm 0.009}$ \\
    \midrule
    \multirow{4}{*}{DS4C}  & 2-Step & $0.803_{\pm 0.000}$ & $0.101_{\pm 0.000}$ & $0.703_{\pm 0.000}$ & $0.008_{\pm 0.004}$\\
    & GS & $0.494_{\pm 0.000}$ & $0.193_{\pm 0.000}$ & $0.301_{\pm 0.000}$ & $1.543_{\pm 0.082}$\\
    & BO & $0.510_{\pm 0.004}$ & $0.148_{\pm 0.022}$ & $0.362_{\pm 0.018}$ & $1.611_{\pm 0.066}$\\
    & \method{} & $0.547_{\pm 0.000}$ & $0.216_{\pm 0.000}$ & $0.331_{\pm 0.000}$ & $0.115_{\pm 0.001}$\\
    \midrule
    \multirow{4}{*}{Contracts} & 2-Step & $0.916_{\pm 0.000}$ & $0.103_{\pm 0.000}$ & $0.814_{\pm 0.000}$ & $0.019_{\pm 0.001}$\\
    & GS & $0.644_{\pm 0.000}$ & $0.268_{\pm 0.000}$ & $0.375_{\pm 0.000}$ & $1.02_{\pm 0.023}$\\
    & BO & $0.650_{\pm 0.020}$ & $0.273_{\pm 0.013}$ & $0.378_{\pm 0.007}$ & $6.905_{\pm 0.536}$\\
    & \method{} & $0.644_{\pm 0.000}$ & $0.268_{\pm 0.000}$ & $0.375_{\pm 0.000}$ & $0.197_{\pm 0.015}$\\
    \bottomrule
  \end{tabular}
\end{table*}

\section{Other Decision Tree Training}

Continuing from the Experiments section, we evaluate \method{} when using a recent decision tree method called Explainable $k$-means clustering~\cite{DBLP:conf/icml/MoshkovitzDRF20}. For this method, we set $\alpha=0$ and fix $k$ using the elbow method. For \method{}, we set $\lambda=1$ and $\epsilon_b = 0.05$ as default values. We repeat each experiment 10 times. In Table~\ref{tbl:accuracyexplainability3}, we observe that \method{} finds solutions with smaller $D+\lambda N$ in all three datasets because complements Explainable $k$-means clustering by optimizing on multiple reference clusterings to minimize the $D+\lambda N$ objective.


\begin{table*}[h]
  \caption{Comparison of \method{} and Explainable $k$-means clustering~\cite{DBLP:conf/icml/MoshkovitzDRF20}.}
  \label{tbl:accuracyexplainability3}
  \centering
  \begin{tabular}{ccc} 
    \toprule
     Dataset & Method & $D+ \lambda N$ \\
    \midrule
    Credit& Explainable $k$-means clustering & $0.950_{\pm 0.020}$\\
    Card & \method{} & $0.528_{\pm 0.006}$\\
    \midrule
    \multirow{2}{*}{DS4C}  & Explainable $k$-means clustering  & $0.654_{\pm 0.020}$\\
    & \method{} & $0.302_{\pm 0.000}$\\
    \midrule
    \multirow{2}{*}{Contracts} & Explainable $k$-means clustering  & $0.950_{\pm 0.029}$\\
    & \method{} & $0.498_{\pm 0.000}$\\
    \bottomrule
  \end{tabular}
\end{table*}

\section{Other Distance Measures for Clustering}

Continuing from the Experiments section, we repeat the monotonicity experiments using other distance measures. We replace DTW distance with Euclidean distance and Jaccard distance with Cosine distance. For \method{}, we set $\lambda=1$ and $\epsilon_b = 0.05$ as default values. We repeat each experiment 10 times. Figures \ref{fig:monotonic_l2} and \ref{fig:monotonic_cos} show that the monotonicity properties mostly hold.

\begin{figure}[h]

\centering
    \subfloat[$k$ versus $D$]{
        {\includegraphics[scale=0.225]{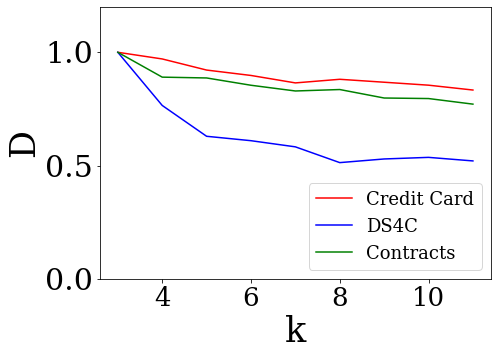}}
        \label{fig:kv_l2}
        }
    \subfloat[$k$ versus $N$]{
        {\includegraphics[scale=0.225]{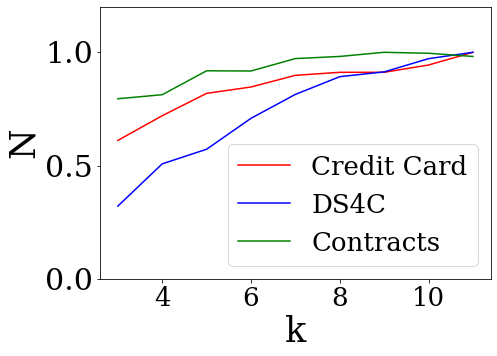}}
        \label{fig:kn_l2}
        }\\
    \subfloat[$\alpha$ versus $D$]{
        {\includegraphics[scale=0.225]{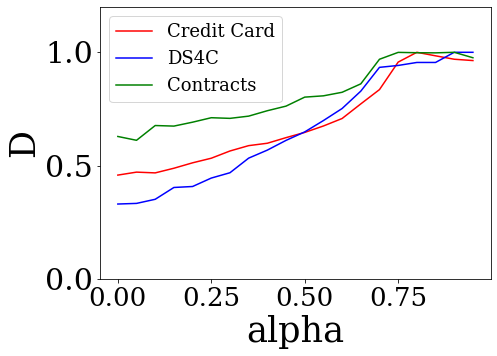}}
        \label{fig:ad_l2}
        }
    \subfloat[$\alpha$ versus $N$]{
        {\includegraphics[scale=0.225]{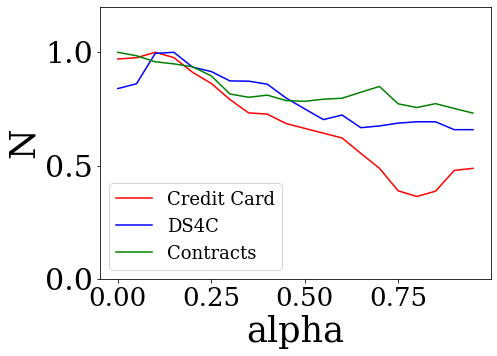}}
        \label{fig:an_l2}
        }
    \caption{Monotonic relationships between $k$ and $\alpha$, and the normalized objectives $D$ and $N$ using Euclidean distance instead of DTW distance.}
\label{fig:monotonic_l2}
\end{figure}

\begin{figure}[h]
\centering
    \subfloat[$k$ versus $D$]{
        {\includegraphics[scale=0.225]{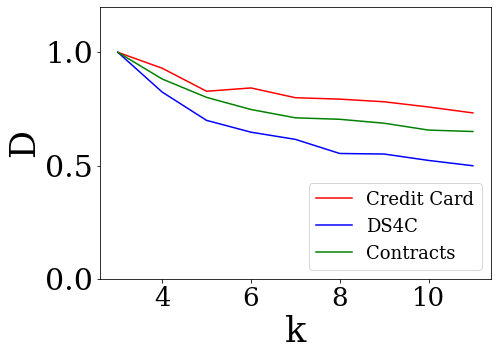}}
        \label{fig:kv_cos}
        }
    \subfloat[$k$ versus $N$]{
        {\includegraphics[scale=0.225]{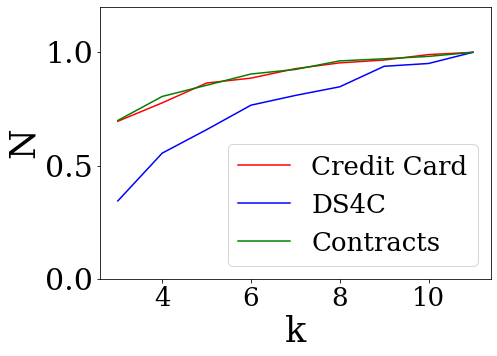}}
        \label{fig:kn_cos}
        }\\
    \subfloat[$\alpha$ versus $D$]{
        {\includegraphics[scale=0.225]{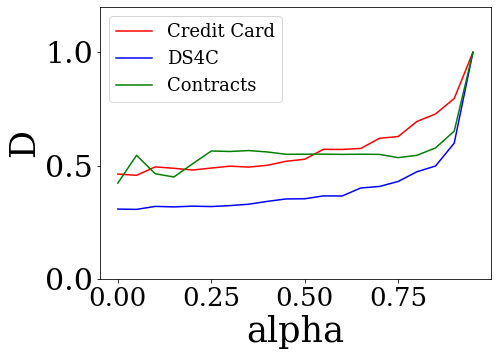}}
        \label{fig:ad_cos}
        }
    \subfloat[$\alpha$ versus $N$]{
        {\includegraphics[scale=0.225]{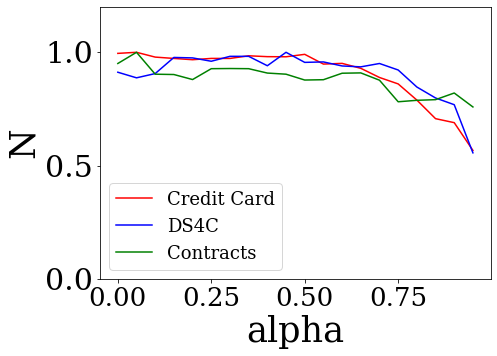}}
        \label{fig:an_cos}
        }
    \caption{Monotonic relationships between $k$ and $\alpha$, and the normalized objectives $D$ and $N$ using Cosine distance instead of Jaccard distance.}
\label{fig:monotonic_cos}
\end{figure}

\clearpage
\section{Decision trees and trends}
Continuing from the Experiments section, Figure~\ref{fig:tree1_og} and ~\ref{fig:tree2_og} show the original images of the decision trees and trends of Figure~\ref{fig:treeandclusters}.
\begin{figure}[H]
\centering
    \begin{subfigure}{0.8\columnwidth}
    \centering
    \includegraphics[width=0.8\columnwidth,trim=0cm 0.3cm 0cm 0cm]{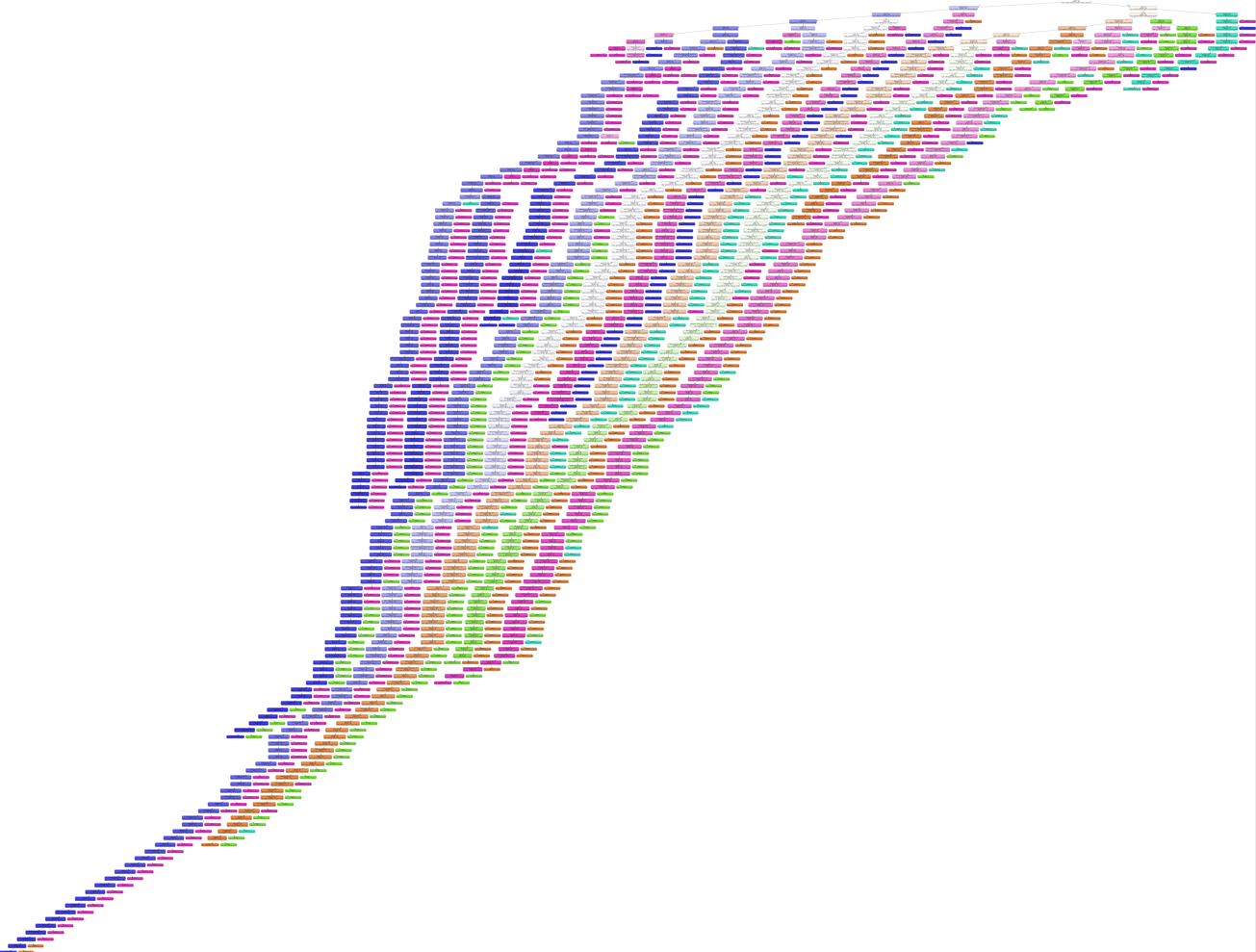}
    \vspace{-0.2cm}
    \caption{Decision tree}
    \label{fig:tree1}
    \end{subfigure}
    \\
    \begin{subfigure}{0.32\columnwidth}
    \centering
    \includegraphics[width=1\columnwidth,trim=0cm 0.3cm 0cm 0cm]{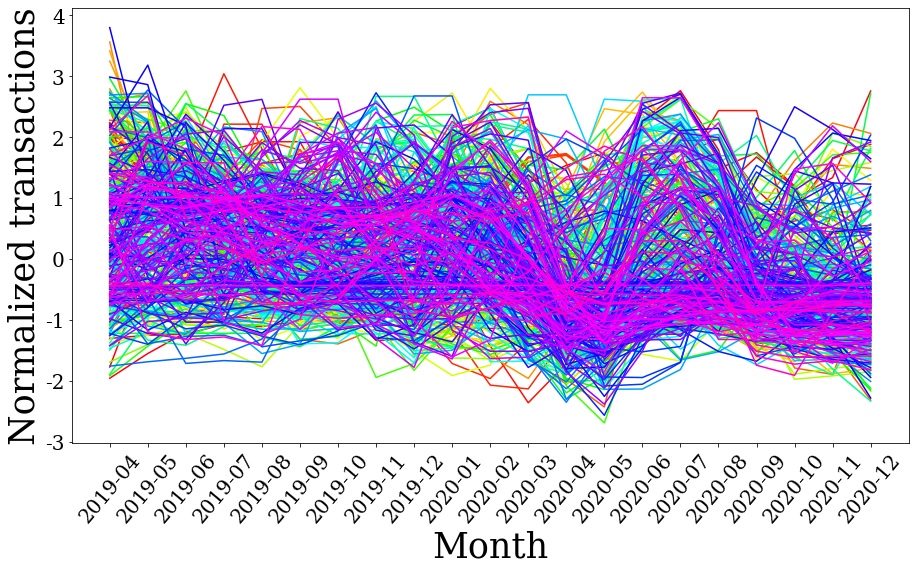}
    \caption{Trend 1.}
    \label{fig:trend0}
    \end{subfigure}
    \begin{subfigure}{0.32\columnwidth}
    \centering
    \includegraphics[width=1\columnwidth,trim=0cm 0.3cm 0cm 0cm]{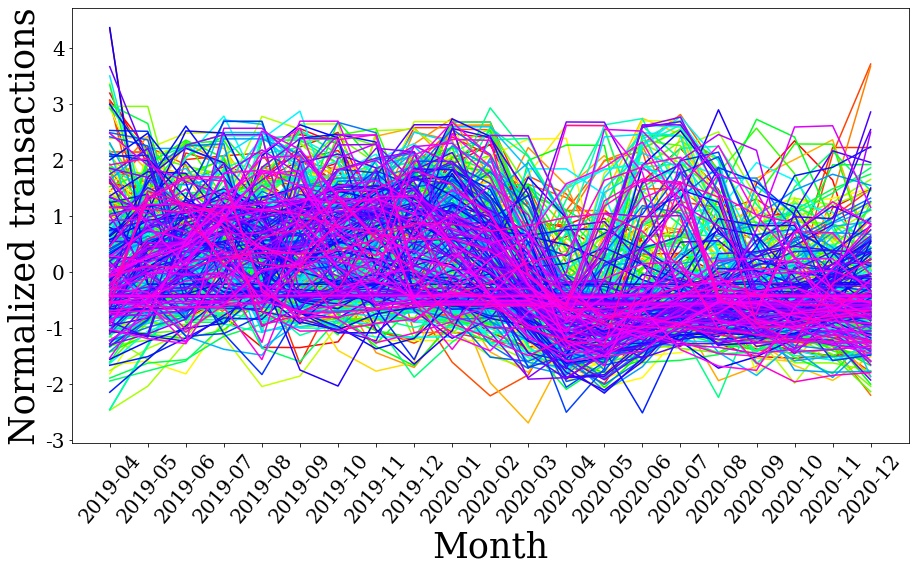}
    \caption{Trend 2.}
    \label{fig:trend1}
    \end{subfigure}
    \begin{subfigure}{0.32\columnwidth}
    \centering
    \includegraphics[width=1\columnwidth,trim=0cm 0.3cm 0cm 0cm]{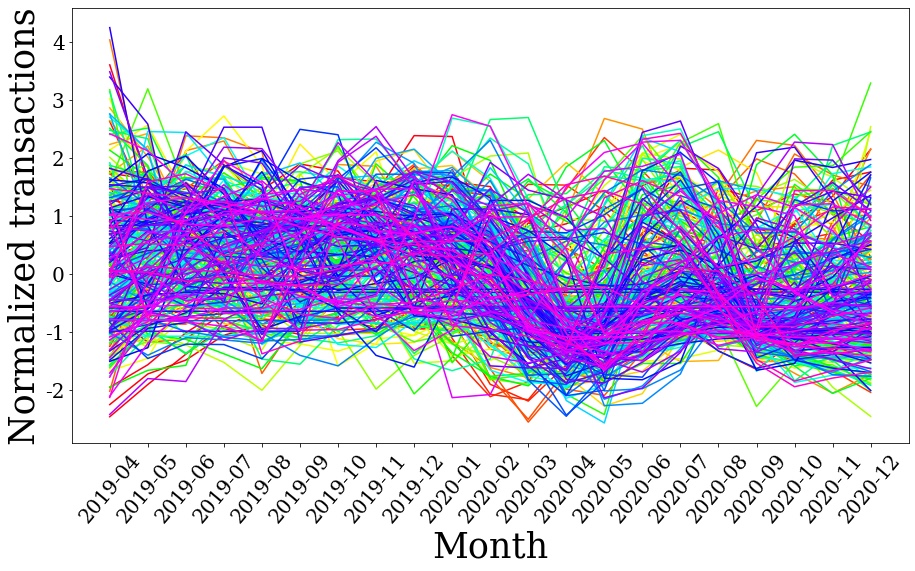}
    \caption{Trend 3.}
    \label{fig:trend2}
    \end{subfigure}
    \begin{subfigure}{0.32\columnwidth}
    \centering
    \includegraphics[width=1\columnwidth,trim=0cm 0.3cm 0cm 0cm]{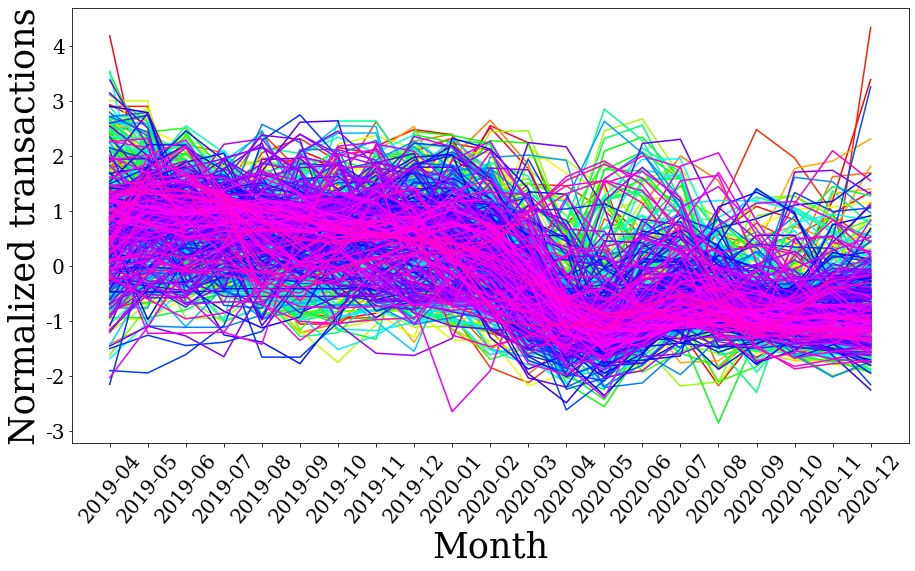}
    \caption{Trend 4.}
    \label{fig:trend3}
    \end{subfigure}
    \begin{subfigure}{0.32\columnwidth}
    \centering
    \includegraphics[width=1\columnwidth,trim=0cm 0.3cm 0cm 0cm]{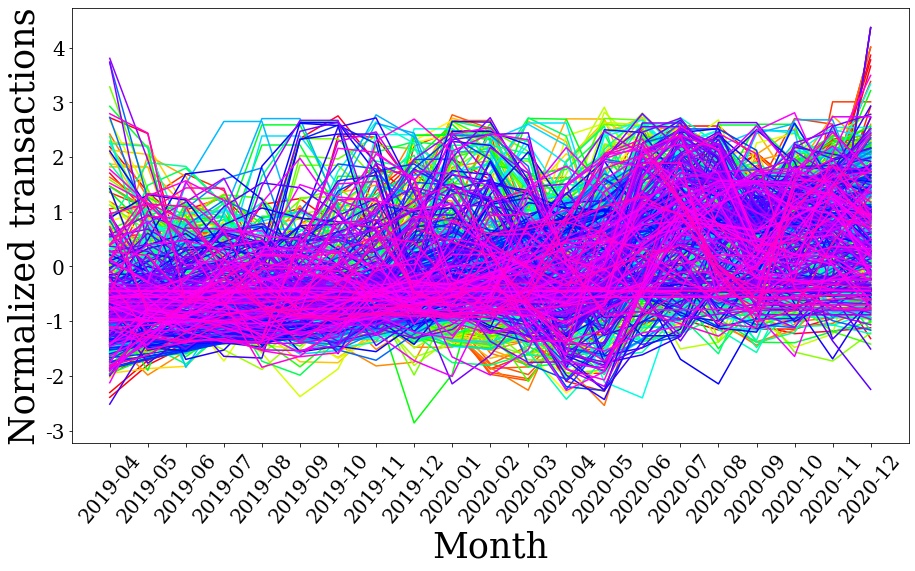}
    \caption{Trend 5.}
    \label{fig:trend4}
    \end{subfigure}
    \centering
    \captionsetup{justification=centering}
    \caption{Tree and trends for the Credit Card dataset for $k=5$ and $\alpha=0.45$. The decision tree has 1,703 nodes.}
\label{fig:tree1_og}
\end{figure}

\begin{figure}[H]
\centering
    \begin{subfigure}{0.8\columnwidth}
    \centering
    \includegraphics[width=0.8\columnwidth,trim=0cm 0.3cm 0cm 0cm]{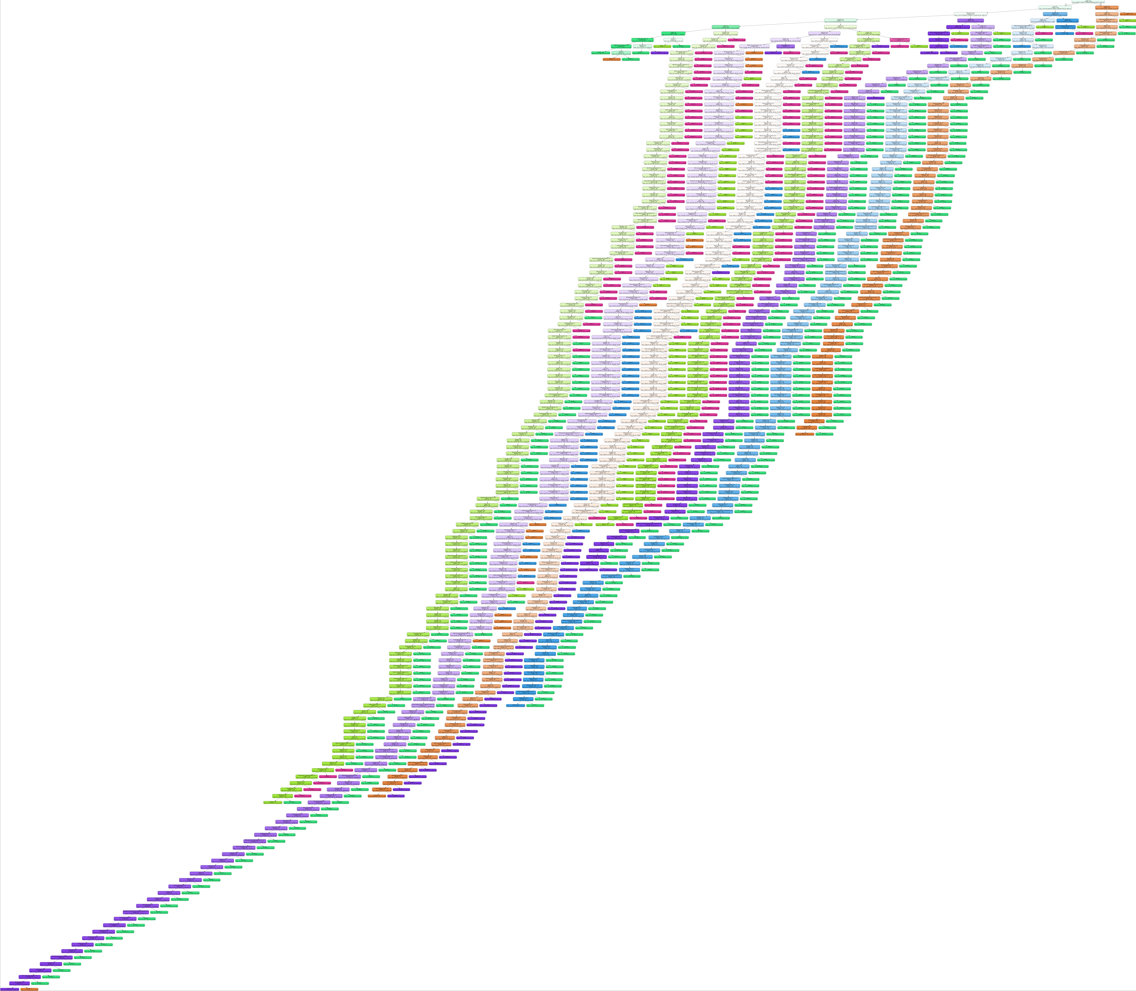}
    \vspace{-0.2cm}
    \caption{Decision tree}
    \label{fig:tree1}
    \end{subfigure}
    \\
    \begin{subfigure}{0.32\columnwidth}
    \centering
    \includegraphics[width=1\columnwidth,trim=0cm 0.3cm 0cm 0cm]{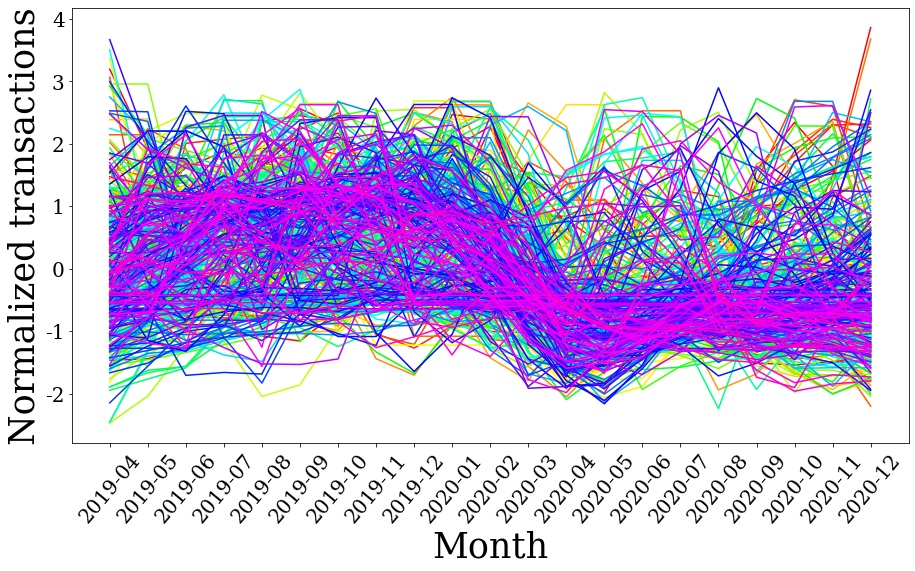}
    \caption{Trend 1.}
    \label{fig:trend0}
    \end{subfigure}
    \begin{subfigure}{0.32\columnwidth}
    \centering
    \includegraphics[width=1\columnwidth,trim=0cm 0.3cm 0cm 0cm]{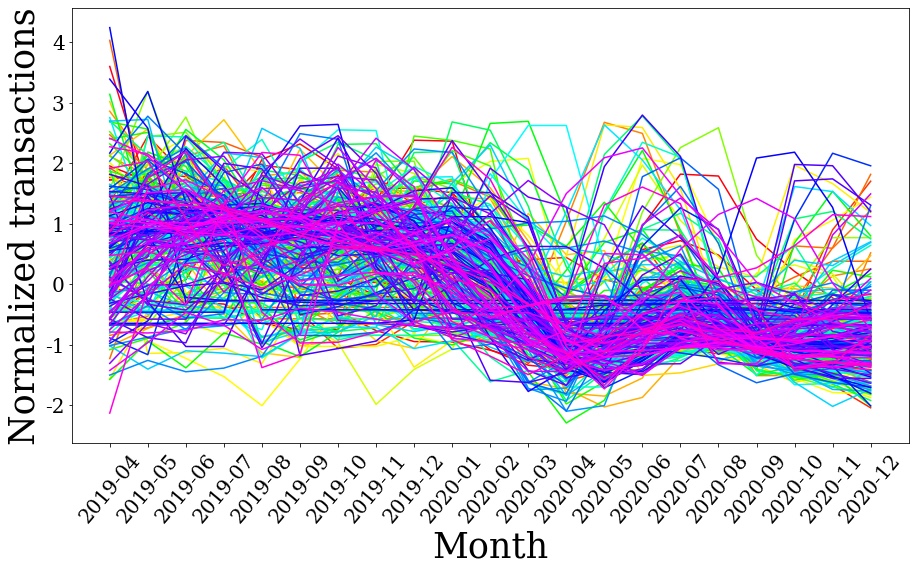}
    \caption{Trend 2.}
    \label{fig:trend1}
    \end{subfigure}
    \begin{subfigure}{0.32\columnwidth}
    \centering
    \includegraphics[width=1\columnwidth,trim=0cm 0.3cm 0cm 0cm]{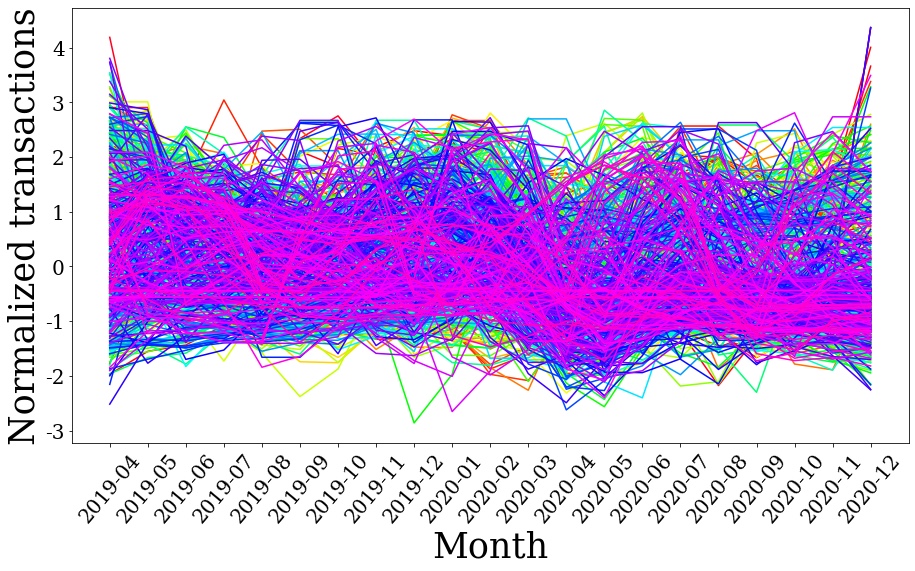}
    \caption{Trend 3.}
    \label{fig:trend2}
    \end{subfigure}
    \begin{subfigure}{0.32\columnwidth}
    \centering
    \includegraphics[width=1\columnwidth,trim=0cm 0.3cm 0cm 0cm]{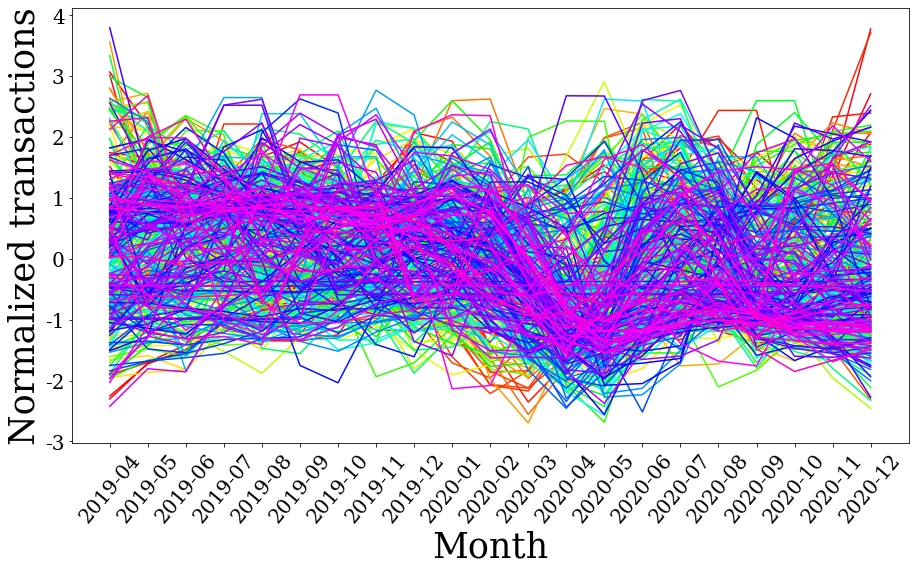}
    \caption{Trend 4.}
    \label{fig:trend3}
    \end{subfigure}
    \begin{subfigure}{0.32\columnwidth}
    \centering
    \includegraphics[width=1\columnwidth,trim=0cm 0.3cm 0cm 0cm]{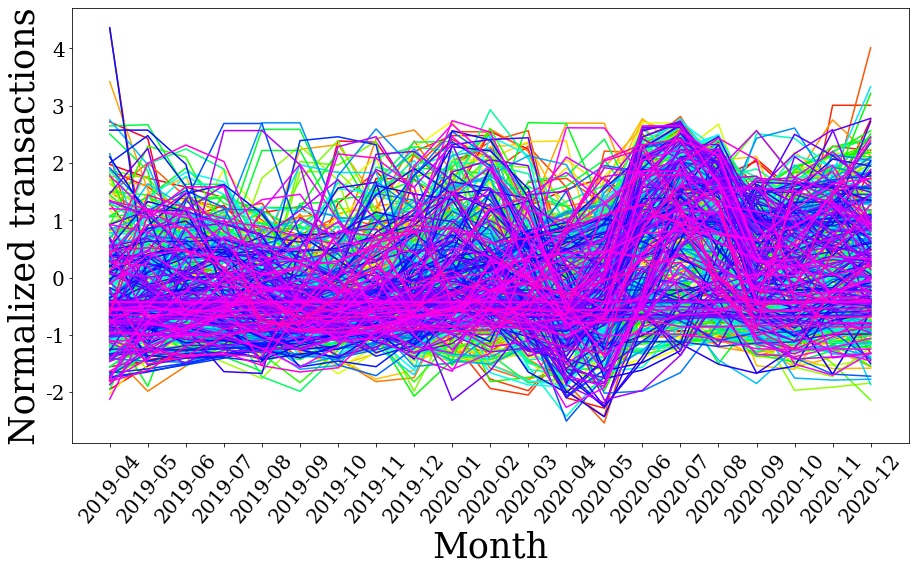}
    \caption{Trend 5.}
    \label{fig:trend4}
    \end{subfigure}
    \begin{subfigure}{0.32\columnwidth}
    \centering
    \includegraphics[width=1\columnwidth,trim=0cm 0.3cm 0cm 0cm]{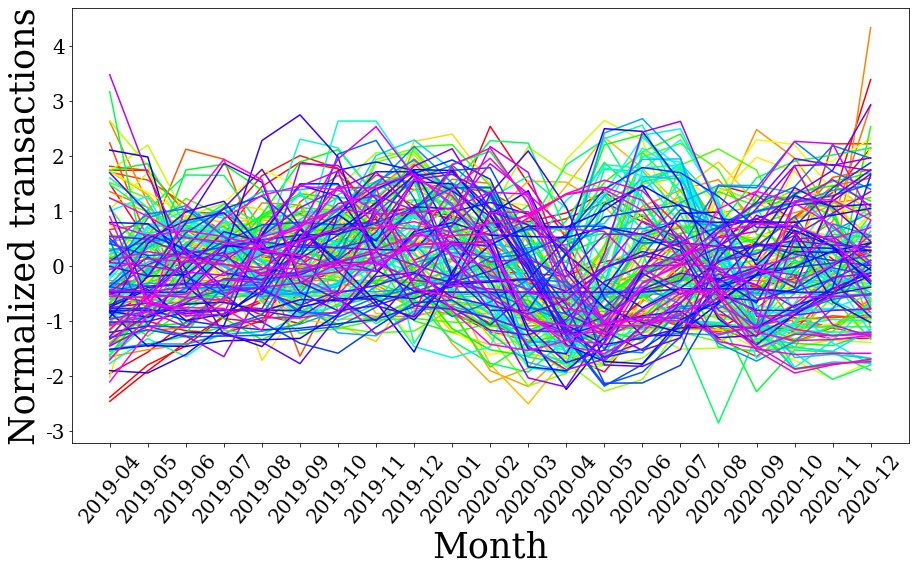}
    \caption{Trend 5.}
    \label{fig:trend4}
    \end{subfigure}
    \centering
    \captionsetup{justification=centering}
    \caption{Tree and trends for the Credit Card dataset for $k=6$ and $\alpha=0.75$. The decision tree has 1,477 nodes.}
\label{fig:tree2_og}
\end{figure}
}{}

\end{document}